\documentclass{article}

\usepackage{PRIMEarxiv}

\usepackage[utf8]{inputenc} % allow utf-8 input
\usepackage[T1]{fontenc}    % use 8-bit T1 fonts
\usepackage{hyperref}       % hyperlinks
\usepackage{url}            % simple URL typesetting
\usepackage{booktabs}       % professional-quality tables
\usepackage{amsfonts}       % blackboard math symbols
\usepackage{nicefrac}       % compact symbols for 1/2, etc.
\usepackage{microtype}      % microtypography
\usepackage{lipsum}
\usepackage{fancyhdr}       % header
\usepackage{graphicx}       % graphics
\usepackage{amsmath}
\usepackage{xspace}
\graphicspath{{media/}}     % organize your images and other figures under media/ folder
\usepackage{epstopdf}
\usepackage[pdf]{pstricks}
\usepackage{graphicx, caption, subcaption}
\usepackage{tikz-cd}

\DeclareGraphicsExtensions{.eps}
\usepackage{multirow}
\usepackage{booktabs}

\usepackage{amsmath}

\title{YZR-net : Self-supervised Hidden representations Invariant to Transformations for profanity detection}

\author{
  Vedant Sandeep Joshi, Sivanagaraja Tatinati \\
  Vedantu Innovations Pvt. Ltd., \\
  Bangalore, India \\
  \texttt{\{vedant.joshi, tatinati.sivanagaraja\}@vedantu.com} \\
    \And
  Yubo Wang \\
  School of Life Science and Technology, \\
  Xidian University,  \\
  Xi-an, China\\
  \texttt{ybwang@xidian.edu.cn} \\
}

\begin{document}
\maketitle

\begin{abstract}
On current {\it e-}learning platforms, live classes are an important tool that provides students with an opportunity to get more involved while learning new concepts. In such classes, the element of interaction with teachers and fellow peers helps in removing learning silos and gives each student a chance to experience some aspects relevant to offline learning in this era of virtual classes. One common way of interaction in a class is through the chats / messaging framework, where the teacher can broadcast messages as well as get instant feedback from the students in the live class. This freedom of interaction is a crucial aspect for any student's learning growth but misuse of it can have serious repercussions. Some miscreants use this framework to send profane messages which can have a negative impact on other students as well as the teacher of the class. These rare but high impact situations obviate the need for automatic detection mechanisms that prevent the posting of such chats on any platform. In this work we develop YZR-Net which is a self-supervised framework that is able to robustly detect profane words used in a chat even if the student tries to add clever modifications to fool the system. The matching mechanism on token / word level allows us to maintain a compact as well as dynamic profane vocabulary which can be updated without retraining the underlying model. Our profanity detection framework is language independent and can handle abuses in both English as well as its transliterated counterpart Hinglish (Hindi language words written in English).
\end{abstract}

% keywords can be removed

\section{Introduction}

% why profranity detection is importatnt.... make it as user generated text... 

In the past few years due to the Covid19 pandemic the adoption of {\it e-}learning platforms has increased significantly. The widespread restrictions have forced students to continue their education via online means which causes them to spend a significant amount of their time watching videos and attending classes. This sudden change from offline to online learning has affected a lot of students therefore making an attempt to build systems that can accurately simulate the experience of offline learning can help in smoothing out this drastic transition. Live classes is one such way that gives the students a chance to escape the monotony of watching recorded videos on a daily basis. The interaction aspect of such classes allow the students to clarify small scale doubts instantaneously and at the same time gives teachers the opportunity to compliment the students on good behaviour. All these tiny bits significantly affect the learning outcome for a student by making the course content more interesting and thus improving their overall engagement on the platform. 

In order to mimic this offline style of interaction there can be a multitude of implementations like live polls or quizzes to check whether the student is paying attention, dynamic interactive diagrams that fuel the curiosity of students by giving them a chance to tinker with it, in-session feedback to understand the student's opinions or the in-class chats mechanism between the participants of a given session. Unlike all the other mechanisms, chats are the most open medium of communication and provide the maximum opportunity to interact with each other. This communication channel is quite important when it comes to serving the dynamic needs of each and every student. For e.g. if a student failed to grasp a concept or has a major doubt which needs to be clarified before starting subsequent topics or the teacher wants to make a quick announcement, all these situations can be easily handled through the chats framework. This solution seems to be the silver bullet when it comes to making the students more engaged but their is a flip side. Lack of regulations on this framework can provide the nefarious elements of a class  an opportunity to disturb the decorum of the session. Posting ill intent messages that are directed towards another student, teacher, racial group or gender can foster the feeling of negativity in the mind of the receiver. Even though such situations occur rarely, they must be addressed instantaneously due to the magnitude of their impact on a student's mind.

%%% provide an example with twitter or other models to showcase fail case... 

% Talk about difficuly in solving with regex or any ML models. 

In any chat, profanity could arise from the usage of an emoji, word or group of words that are forbidden by the general ethical rules of society in day to day conversations. Semantically these single or grouped profane words could have relevant or irrelevant meaning but since they are crafted to harm someone's mental stability, removing such chats from the platform becomes imperative. The main challenge lies in effectively separating the profane chats from the non-profane ones because in many cases a slight deviation from the correct spelling can change the intent of the overall message. As mentioned, the percentage of profane chats generated on a daily basis is quite low and therefore generating a data set in order to learn the semantics of profane tokens would be challenging. In this problem, semantics of the words play an important role but it is the syntactical structure of the profane word which is the key element that needs to be learnt. Usually profane / cuss words are used by students to direct some anger or to have some fun in the class. These chats are usually not well structured and are riddled with short forms, undefined lingos and a lot of grammatical errors, which clearly indicates that, in order to detect profanity we need to build mechanisms that are impervious to such mistakes / noises.
Therefore we need to develop a detection system that is :

\begin{itemize}
    \item Invariant to spelling variations and tricks used by students to fool the system.
    \item Accurately detects all edge cases without letting negative messages pass and at the same time not hinder student experience by repeatedly blocking appropriate messages.
    \item Support real time inference for rapid results.
\end{itemize}

There is an added challenge for our platform in this problem statement. Most of the previous works in this domain have given high performance in ill-intent tweets detection but their ability is only limited to English language which has a defined structure and contains a lot of well researched resources. Our platform caters to students from different backgrounds and most of them feel comfortable by communicating in the transliterated language Hinglish \cite{HINGLISH}. Hinglish is not a language per se, but based on phonetic similarity between sounds of various letters or words, students are able to write Hindi in English characters (to slightly ease the problem we have disabled the support to post chats in Devnagiri script). Since Hinglish is not a well defined language, there are no grammatical, syntactical or spelling rules therefore it primarily depends on sound similarity which posses a big challenge for deep learning models to perform pattern searching and learn differences between tokens with the same spelling but different meanings due to pronunciation. Current NLP models such as LSTM \cite{LSTM}, GRU \cite{GRU}, transformer \cite{TRANSFORMER} etc. all depend on a fixed vocabulary to get a vector representations of each token in a string. As mentioned Hinglish lacks a well defined vocabulary therefore hard coding every possibility of a given word would be infeasible.

In this work, we propose YZR-Net (character level), robust string matching model which is invariant to the possible tricks used by the students to fool the system. The goal is to focus on the character level structure of key profane words instead of trying to learn their semantics.
For our intended audience i.e. the students, we formulate the problem such that, if there is a single profane word in the chat, we mark it as an ill-intent one thereby preventing it to be published on our platform. Inspired by the recent self-supervised work in \cite{looking}, our approach is fully data-driven in that it does not require any manually generated labels. Trying to manually find profane entries in a huge stack of chats where majority of them are non profane in nature is a humongous task. This self supervised approach allows us to maintain a limited size dictionary and supports dynamic addition of more profane tokens without re-training the underlying model. Our work is an adaptation of SimCLR \cite{SIMCLR} for a NLP problem statement. This SSL methodology trains a base encoder network in a Siamese approach and generates positive pairs by augmentation policies that mimic the tricks utilized by students and negative pairs by contrasting with other elements in a batch. Our model is able to learn a latent space where the key tokens of our profane dictionary form anchor points and all its noisy variations ( spelling differences and self censoring ) are placed in close proximity as visualised by 2D UMAP projections \cite{UMAP}. The model is trained on character level embedding which allows it to focus on fine grained bi,tri-gram level features that allow it to place profane tokens close to the anchor points even if there are spelling variations.

This paper is organized as follows: we discuss the related work in Section 2, specify the task definition and describe our
self-supervised learning based models in Section 3. In Section 4, we explain the experimental setup, in Section 5 we
provided the implementation details. Lastly, we present the evaluation results as well as a detailed analysis.

\section{Related Works}

\subsection{Traditional Methods}
Packages such as go-away \cite{GOAWAY} make use of regular expression to capture different possible variations of profanity that a student might enter to fool the detection system but this hand-engineered method would always fail in case of misspelled profane tokens \& the overall detection system would not be able to stop the negative intent of the message being typed on our platform. Building a dictionary of all possible spellings for a given token is an activity which is both resource as well as time intensive \& would yield only a marginal improvement in performance for the overall detection system. 

Burnap et al. in \cite{BURNAP} made use of twitter for generating a data set on hate-speech detection. They performed the exercise of scrapping right after the occurrence of an event that had the potential to trigger a lot of cyber-hate. The authors trained an ensemble of probabilistic, rule-based \& spatial classifiers on features that were hand-engineered to focus on grammatical as well as semantic contents in a given tweet. To further boost the performance the authors tried an experimental lexical parser that generated a typed dependency list which highlighted the grammatical relationships in a sentence. The final goal of the study was to use these predictions in a statistical model to forecast the likely spread of cyber hate.

Davidson et al. in \cite{DAVIDSON} explained the difficulty in defining the boundary between hate \& non hate speeches on online platforms by emphasizing on the low precision \& high recall performance of traditional lexical based methods. The authors build manual features that contain n-grams of characters along with a sentiment score for each tweet. The authors trained a linear SVM on the inputs whose dimensionality is reduced by logistic regression to get a F1 score of 90 percent for the task of hate, offensive \& neither class classification.

\subsection{Deep Learning Methods For English}

Zhou et al. in \cite{ZHOU} highlight the problem of representational sparsity \& high dimensionality with traditional bag of words method \& tackle the problem of profanity detection by learning low dimensional representations of hateful comments using paragraph2vec  \cite{PARA2VEC}. The authors use an updated objective to learn a more structured space for joint modelling of words \& comments. Training binary classifiers on this learnt latent space lead to an improvement of 11\% AUC score when compared to binary classifiers trained on traditional encoding schemes.

Zang \& Luo in \cite{LUO} explain the challenges to extract robust as well as unique features to identify hate speech on online platforms thereby highlighting the limitations of hand engineered representational methods. The authors implement a base CNN encoder which is applied in a skipped manner that allowed it to extract n-gram like features from the raw input text. The encoder features are fed into a GRU \cite{GRU} time step wise followed by a global max pooling layer before applying a softmax for multi-class classification. This supervised setup led to an improve of 8\% of F1 score when compared to the benchmark for hate speech detection.

Cranefield et al. in \cite{CRS} perform the task of profanity detection for software engineering communities like github \& stack-overflow along with a conflict reduction system (CRS) that identified the offence \& recommended the steps to minimize it. The authors manually annotated the comments from these platforms along with text augmentations to better train the deep learning models. The authors utilized pre-trained BERT \cite{BERT} on wikipedia corpus for the task of comments classification \& achieved a prediction accuracy of 97\%. For the CRS system the authors further classify the hateful comments into finer labels along with a regex component that specifically identified the hateful content.

Kapoor et al. in \cite{MIND} solve the problem of profanity detection for tweets in the code switched language Hinglish. They create a fixed vocabulary from the HOET  in \cite{HOET} and hate speech detection dataset in \cite{DAVIDSON} to learn word embedding via the GLoVE algorithm in \cite{GLOVE}. These embeddings serve as an initial distributed representation of tweets which is further pre-trained on the Davidson dataset. The authors use an LSTM pipeline with dense layers to perform 3 class classification along with an aggressive dropout layer before the input to the LSTM. After pre-training the authors use transliteration and translation dictionaries to map common Hindi words and their variations to English words so that in the final fine tuning step on HOET the model is able to better use the pretrained weights. They achieve a peak accuracy of 87\% on HOET dataset.

\section{Profanity Detection Algorithm}

We denote the chats database generated by the students on our platform by $ \mathbf{\chi} = \{{\bf x}_1, {\bf x}_2, \cdots, {\bf x}_N\} $, where ${\bf x}_i$ represents $i-$th chat. Each chat $x_{i} = [c_{1}, c_{2}, ..., c_{m_{i}}]$, is a string which is nothing but a sequence of characters \& $m_{i}$ denotes the variable length of each chat. In order to ease the problem of profanity detection we have restricted the input character set $\psi_{domain} = \{$ 26 letters of English alphabet, 3 special characters \& 1 out of vocabulary token $\}$ . The definition of $\psi_{domain} $ is imperative to our problem statement because our model can detect profanity only in English \& the code switched language Hinglish ( Hindi in English ). 

The objective is to learn an encoder $f(.)$ which is able to build a structured latent space where a given token \& all its variants such as short forms, misspelled versions,  phonetically similar sounding cases etc. are placed close to each other. In a given chat ${\bf x}_i$ a token is defined as a contiguous sequence of characters which is preceded as well as succeeded by the space character i.e. space based tokenization. A list of tokens in a chat can be denoted by $x_{i} = [(c_{1}, c_{2}, c_{3}), (c_{5}, c_{6})..... (c_{a}, ... c_{b})]$ where the missing character indices denote the space character that $\in \psi_{domain}$. The model is able to learn such a space by minimizing the following objective :

\begin{equation*}
    L (e_i,e_j) = -log( \frac {exp^{ sim(e_i, e_j)/ \tau} } { \sum_{k=1}^{2N} 1_{[k \ne i]} exp^{sim(e_i, e_k)/ \tau} } )
    % \label{ntXentLoss}
\end{equation*}

where $e_{i} = f((c_{a}, ... c_{b}))$ is the embedding generated by the model for a token in the chat $x_{i}$. The sim(.,.) function in the given loss objective is cosine similarity that measures the closeness of 2 embedding in the latent space \& $\tau$ is the temperature term that controls proximity of any pair of points. The goal is to build a robust matching module which is able to capture most of the possible variations for each profane token. This allowed us to maintain a compact dictionary of key profane tokens only \& also add newly found profanity without retraining the whole pipeline. The algorithm contains 3 main parts : a) Pre-processing \& Normalisation module, b) Cleaning \& Dictionary match module, c) Latent space search module.

\subsection{Pre-processing \& Normalisation Module}
The raw data generated from chats contains a lot of variations which is the primary reason why a simple regular expression based matching pipeline fails to eradicate the profanity problem completely. Special symbols, character accents, self censoring, random use of digits etc. are some of the possible ways where a student can fool a simple string matching module to convey the profanity in the chat.

In order to ease the task of learning by normalizing all input characters to the same scale, following steps are implemented in this module : 
\begin{itemize}
    \item Removal of all email handles, url links \& student names from the chat.
    \item Mapping a subset of special characters to their English character counterparts. E.g. \$ -> s
    \item Removing all numbers.
    \item Accent normalisation in chats.
    \item Removal of emojis.
    \item Space normalisation \& removal of all mathematical operator symbols.
    \item Mapping the remaining special characters to a single "*" character.
    \item Removing cases where a single character is consecutively repeated 3 or more times.
    \item Lower casing all input characters.
\end{itemize}

All these functions have been designed based on the nature of the our chats database \& can be easily modified for data sources with different nature. These functions are necessary for the forthcoming modules to have better F1 scores in testing conditions.

\subsection{Cleaning \& Dictionary Match Module}

Once the previous module reduces the variance in the input chat we perform space based tokenization to get a list of suspicious tokens. The main motivation behind building this list of suspicion is that on a daily basis a lot of chats are generated on our platform \& a great deal of them are not profane in nature. Passing every token from each chat through the model \& performing a search in latent space would be too time consuming therefore we decided to remove all the tokens that we were confident of not finding in any form of profanity. These high confidence, non-profane tokens are maintained as dictionaries for both English as well as Hinglish. To further reduce false positives we maintain an extra dictionary for certain key terms that are specific to our platform. While performing this cleaning step we also compare the tokens with profane vocabulary in order to find direct matches \& reduce the overall search time in the pipeline. This direct matching mechanism allows our model to build on top of the baseline accuracy achieved by simple regex.

In NLP space based tokenization is one of the most easiest ways to break a given string into independent entities but the biggest problem with this approach for our chats dataset is that if the student delimits the profane word with spaces then this approach of tokenization will lead to entire loss of profanity context and the pipeline will never be able to predict the chat as profane.

\begin{equation*}
    Input \: chat : a \: b \: u \: s \: e  => [a,b,u,s,e] \: list \: of \: tokens
\end{equation*}

In order to tackle this problem we use the following notion, if we have 2 consecutive suspicious tokens and the current suspicious token has length less than or equal to 4 characters then we concatenate it with the previous one in order to recover the original profanity context. The upper limit of 4 characters for suffix concatenation is kept so that our model does not get confused if the profane token is followed by some gibberish in the chat. If multiple, consecutive suspicious tokens are concatenated then it will attenuate the signal of the main profane token resulting in  reduction of similarity scores with the keys in profane dictionary thereby making it difficult for our model to make a reliable match.

\subsection{Latent Space Search Module}

\begin{figure}[h]
    \centering
    \includegraphics[width=3.5 in]{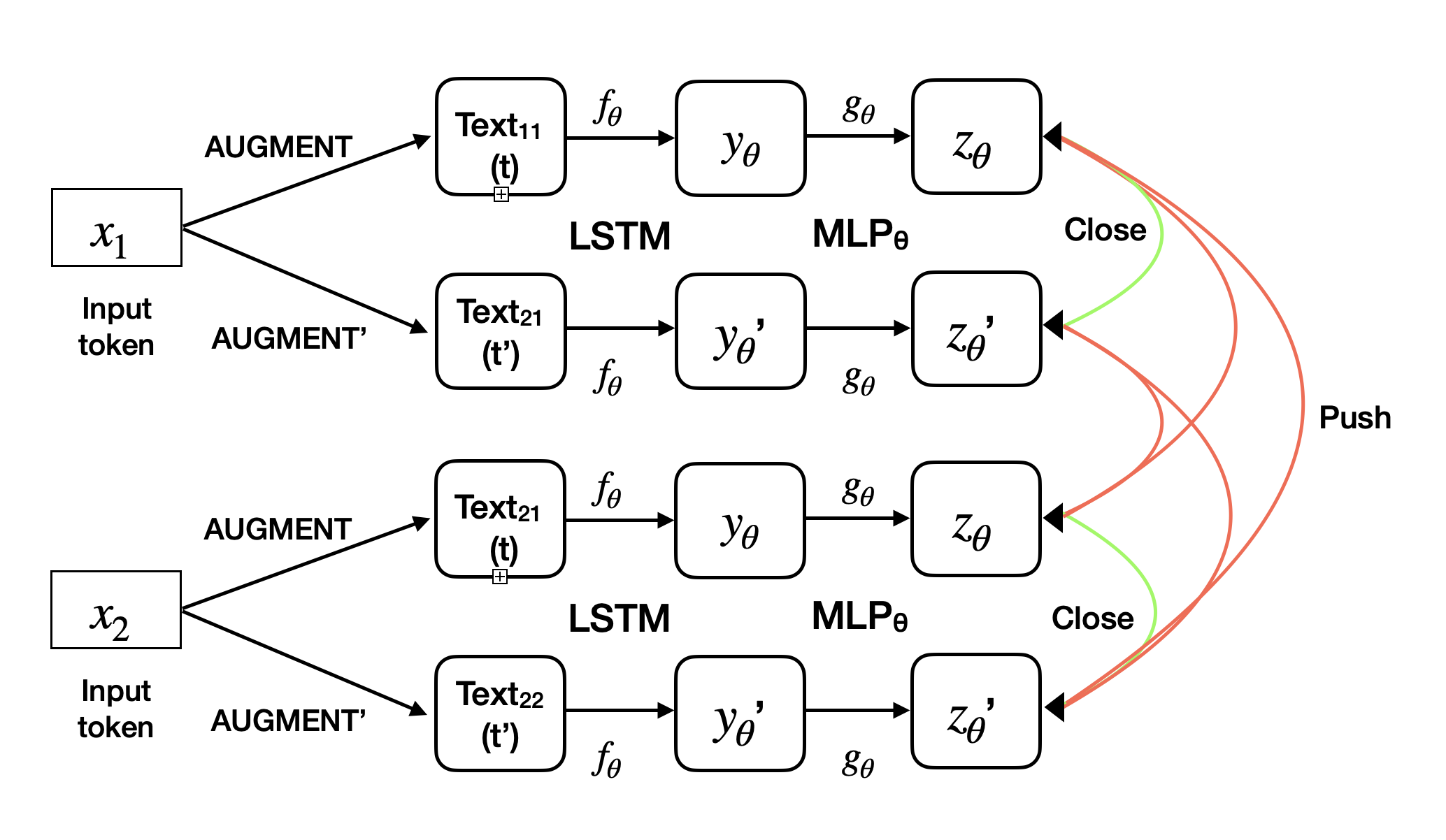}
    \caption{Model Architecture}
    \label{model_arch}
\end{figure}

The goal of our model is to learn a structured, lower dimensional latent space where it is able to place different variations ( noisy versions ) of a given token in dictionary close to each other. The main objective of our pipeline is single token profanity detection by performing robust matching to keys in dictionary. The architecture of our proposed solution is depicted in Fig. \ref{model_arch}. 

The architecture is made up of 2 arms that contain identically parameterized encoder \& projector networks. Given an input token $x$, where each of its constituent characters  $c_{i} \in \psi_{domain}$, we generate 2 variants $t$ \& $t'$ by the augmentation functions which are sampled from a parametric space where we either randomly delete certain characters from the token or replace them with the self-censoring character : "*" that  $\in \psi_{domain}$. We have also added a upper bound limit for these augmentations based on the length of the original token, so that aggressive augmentations do not drop too much relevant signals thereby incentivizing the model to perform false matches.
The pair generation step is given by :

\begin{equation*}
    t = AUGMENT(x) \: \& \: t' = AUGMENT'(x)
\end{equation*}

The encoder network is a 3 layer-stacked LSTM model where the word / token encoding vector is the hidden state in the last time step of the final layer of the model. The LSTM network is a character level model which is trained on the vectors generated by the embedding layer whose vocabulary is $\psi_{domain}$. The embedding vector is given by :

\begin{equation*}
    y_{\theta} = h_{N} = LSTM(t_{N}, h_{N-1})
\end{equation*}

N denotes the final time-step. The encoding generated from the previous step is passed through a MLP represented by $g_{\theta}$ to control the dimensions of the vectors that will eventually be used for search in the latent space. The final search vector for a given token $x$  is given by :

\begin{equation*}
    z_{\theta} = g_{\theta}(y_{\theta})
\end{equation*}

Based on the instance discrimination objective, our model is trained with the help of positive as well as negative pairs. The loss function tries to increase the similarity of augmented tokens \& reduce the similarity of different tokens in a batch while training. Unlike the vision domain \cite{SIMCLR}, the unsupervised instance discrimination objective on unlabeled corpus of tokens, works quite well for our use-case because our training dataset is made up of a set of unique tokens therefore we are quite sure that we are contrasting from a correct set of negative points given an anchor point. 

\section{Experimental Setup}
\subsection{Dataset}
As explained in the previous sections, this is a new approach towards profanity detection therefore there are no publicly available datasets for bench-marking. The SSL objective for our model is to perform robust token matching therefore we decided to train our model on all the available words in English \& Hinglish. We also added our profane vocabulary into the training dataset which led to the following training statistics :

\begin{itemize}
    \item English Tokens  : 6675
    \item Hinglish Tokens : 232,917
    \item Profane Tokens  : 654
    \item Total Number of unique tokens : 239,290
\end{itemize}

All the English tokens are taken from the nltk library corpus \& the Hinglish tokens are taken from the work in \cite{MIND}. One of our goals was to train the matching module in such a manner that given a dynamic dictionary, our model will be able to robustly detect all the possible variations of the dictionary keys without any re-training. In order to achieve this goal we have attempted to train our SSL model with as many tokens as possible. This idea allows us to build a latent space where the anchor tokens ( the ones with the correct spellings ) are at sufficient distance to each other in order to avoid any false positive matches.

% \begin{figure}[h]
%     \centering
%     \includegraphics[width=3.5 in]{char_dist.eps}
%     \caption{Training Data Character Distribution}
%     \label{char_bar}
% \end{figure}

\subsection{Performance Metrics}
For our domain, the most important objective is to have lesser false negatives i.e. we reduce the incorrect prediction for ill intent chats so that we can protect our students. Therefore individual metrics such as precision, recall and F1-scores are used in the model comparison baselines. The value of these metrics also give us an idea about how strict or lenient our matching model should be when it comes to decision making threshold while performing cosine similarity with key profane tokens in the vocabulary.

\subsection{Baselines}  
For comparing the performance of YZR-Net with readily available solutions we replicate the work done in \cite{MIND} for profanity detection. The comparison is carried out on the Davidson dataset \cite{DAVIDSON} in order to showcase the downstream performance of YZR-Net without being explicitly trained on the testing dataset. We also use a simple regex comparison, which uses the same vocabulary as YZR-Net in order to showcase the improvement of the SSL approach to capture difficult edge cases and give a overall better score when both the system are used sequentially.

\section{Implementation details}
\subsection{Training Data} 
The dataset is split into train / valid splits based on the following percentages : 70\% / 30\%. The validation set contains samples that have been augmented with less intensity \& therefore the loss computed on this set acts as an indicator for model over-fitting. The idea is that if the model is not able to place the less augmented vocabulary in the validation split closely in the latent space i.e. exhibit high NTXent loss value, then the model is not able to generalize on unseen samples for the task of robust token matching. 

Another restriction we added while implementing the augmentation pipeline was that while corrupting a given token randomly to generate a pair of positive samples, we always ensured that the anchor word's first and last character is always retained. This is done in order to reduce the possible number of false matches (for eg. without this restriction our model was matching the word "itch" with "b*tch" due to high character overlap) and in-case a word is posted on the online chat that lacks a first or last or both the characters, then lack of the context makes it to general to match with any token. Since we are focusing on independent token level prediction  and not using any context from adjacent words, such restrictions become important.

Based on the initial EDA of chats data, we have also limited the length of the longest word in the vocabulary to 24 characters. This restriction was based on the common knowledge that no student would post a word on the chat which is too long and especially for profane tokens. We have set the padding of words to be same as the max length token, this restriction ensures that the model will not start to use padding as a key feature to match the tokens being input in our pipeline.

\subsection{Architecture} 
The SSL models are trained on the instance discrimination objective in \cite{INFONCE}. Our model contains 3 layers of stacked LSTMs \& each layer makes use of the tanh non linear activation function. The output from the final layer of the last time step is fed into a batch normalisation layer with dropout to prevent over-fitting and later it is resized by a ReLU activated dense layer to 64 dimensions to perform rapid search in the final latent space. The extra added dense layer is based on the information drop phenomenon observed in SimCLR \cite{SIMCLR}.  

The NTXent loss function is defined with a low temperature of 0.07 in order to learn token encoding which is densely clustered with anchor points. The automatic hard mining nature of the loss function pushes words with different character distributions away and tries to keep only the corrupted versions of a given close to each other.

In order to perform rapid matching of tokens in the latent space with profane words we employ Hierarchical navigable small world (HNSW) \cite{HNSWLIB} that  builds a multi-layered graph structure where each layer represents a proximity graph which is constructed based on our compressed representations of tokens. Elements are populated in the proximity graph based on their closeness to each other. In the final deployment pipeline we only place the elements of the profane dictionary in this graph structure and try to remove the non-suspicious tokens as much as possible in the initial modules to avoid false positive matches. 

\subsection{Optimization}
The model is trained with Adam optimizer \cite{ADAM} on a learning rate of 1e-04 with cosine decay to ensure proper convergence. Based on the intuitions of \cite{SIMCLR}, we train our SSL model with large batch size of 2048 to generate a lot of negative pairs to contrast from. All our experiments are carried on a cloud system with 16gb P100 GPUs. The models are trained for 3000 epochs \& it took 5 hours to reach a saturating threshold value for the loss function on a validation split. Models with the best validation loss value are saved since they exhibit the best generalization capabilities.

\begin{figure}[h]
    \centering
    \includegraphics[width=4.5 in]{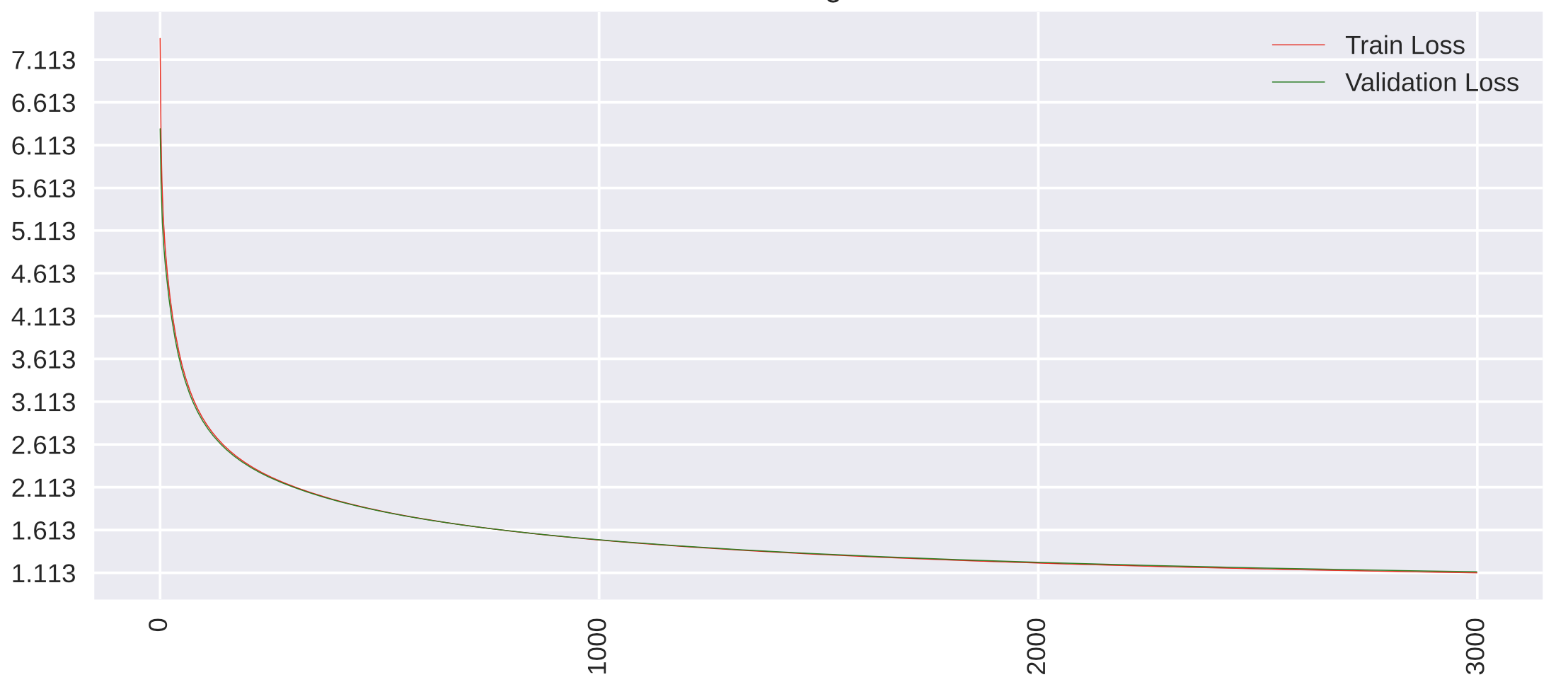}
    \caption{Training Plots}
    \label{char_bar}
\end{figure}

\section{Results and Analysis}

\subsection{Baseline Performance Comparison}
In order to test the performance of token level profanity scheme, we compared YZR-net pipeline with the vanilla regular expression matching (using the same profane vocabulary, cleaning and tokenization pipeline as that of YZR-Net) and the LSTM model introduced in \cite{MIND} for hate speech classification (it uses the pre-processing pipeline as mentioned in the paper). For the aforementioned work, we implement the model from scratch as mentioned in the paper, but during evaluation instead of performing a 3 class classification on hate, offensive or neither speech we combine the 2 classes of hate and offensive into one. This is done so that we can have an apples to apples comparison between our implementation and the closest solution available for our problem.

We compare all the models on the Davidson dataset in \cite{DAVIDSON}, which is a collection of tweets labelled as hate, offensive or neither class. We keep the cosine similarity threshold for detecting a match to profane vocabulary as 0.8 in our YZR-Net pipeline. The results are summarised as follows :

\begin{table}[h]
\centering
\caption{Performance Comparisons}
\label{tab:lakdc_data}
\centering
\begin{tabular}{ lllll }
\hline
 Model Name & Precision & Recall & F1-Score \\
 \hline
 Profane Class & & & \\ 
 \hline 
 Regex & 96\% & 82\% & 88\%\\
 Baseline LSTM & 98\% & 93\% & 96\%\\
 YZR-Net & 98\% & 86\% & 92\%\\ 
 \hline
 Not Profane Class & & & \\ 
 \hline 
 Regex & 55\% & 96\% & 70\%\\
 Baseline LSTM & 73\% & 90\% & 81\%\\
 YZR-Net & 55\% & 96\% & 70\%\\
 \hline 
\end{tabular}
\end{table}

The results clearly show the improvement of in-variance to noise in our model over the vanilla regex pipeline. The modest improvement in F1 score is contributed to the fact that the Davidson dataset contains tweets that have correctly spelled profane words i.e. the ones that have a direct match with our vocabulary. The precision values for Non Profane class takes a massive drop and it is due to the conservative profane vocabulary created by us for our target audience. In order to have a cleaner platform, a lot of words have been kept in our vocabulary and therefore some tweets which might not seems profane to adults but could negatively impact young students are marked as profane by both regex and YZR-Net. Please note YZR-Net was never trained or fine tuned for the the Davidson dataset and it only needs to be pre-trained once on a set of tokens. For deploying YZR-Net in various setups, one only needs to change the target profane vocabulary and the operations to be used in cleaning pipeline.

\subsection{Robust Word Detection}   

The biggest problem with the regular expression mechanism and the general deep learning models in NLP is the fixed vocabulary of tokens. Variations in the spelling  of the profane tokens can be done intentionally or due to lack of rules in the language being used ( major problem of Hinglish ). There have been potential solutions for handling the out of vocabulary issue in deep learning models based on sub-word tokenisation schemes, especially byte level BPE in \cite{BPE} but the fundamental principal on which it is based, is well suited for structured languages such as English and not Hinglish. Even after these variations the general intent of the profane word is retained and regular expression pipelines miss them quite often. The deep learning models map out of vocabulary (OOV) words to a constant embedding which will always result in a constant prediction. Constant mapping to OOV token for a misspelled or self-censored profane word would always be a constant prediction which does not solve our problem of achieving cleaner environment for students and teachers.

YZR-Net on the other hand is trained for solving the aforementioned problem. The main objective of our model is to handle all those cases where the regex pipeline fails. While training on this objective another problem that comes up quite often in deployment is false positive matches. If we train our model to give same / close representation to :

\begin{equation*}
    Profane \: word : f*ck  \: | \: Augmented \: counterparts : f**k, \: fck, \: fk \: etc.
\end{equation*}

Since the model is learning character level embeddings ( to avoid OOV problems ) and making predictions on each token independently, we can have situations where 2 semantically different words having similar characters, matched by our model. For e.g. :

\begin{equation*}
    Profane \: word : co*k  \: | \: Mapped \: to \: : cook
\end{equation*}

The random character deletion augmentation highlighted the key characters of the profane token which led to a proper match with a non-profane token. One solution to avoid such situations is to use a high cosine similarity threshold but that could lead to drop in recall performance of our model which would be detrimental for our problem statement. 

In order to solve this trade-off between precision and recall, we decided to create an extra vocabulary of non-profane words that is implemented in the tokenization module of our pipeline. This vocabulary contains popular Hinglish and English words along with some added lingos/terminologies used on our platform. This cleaning procedure allows us to send lesser tokens to the model inference pipeline thereby saving our compute and time resources along with improved overall F1 scores.

\subsection{2D Latent Space Visualisation}
\begin{figure}[h]
    \centering
    \includegraphics[width=3.0 in]{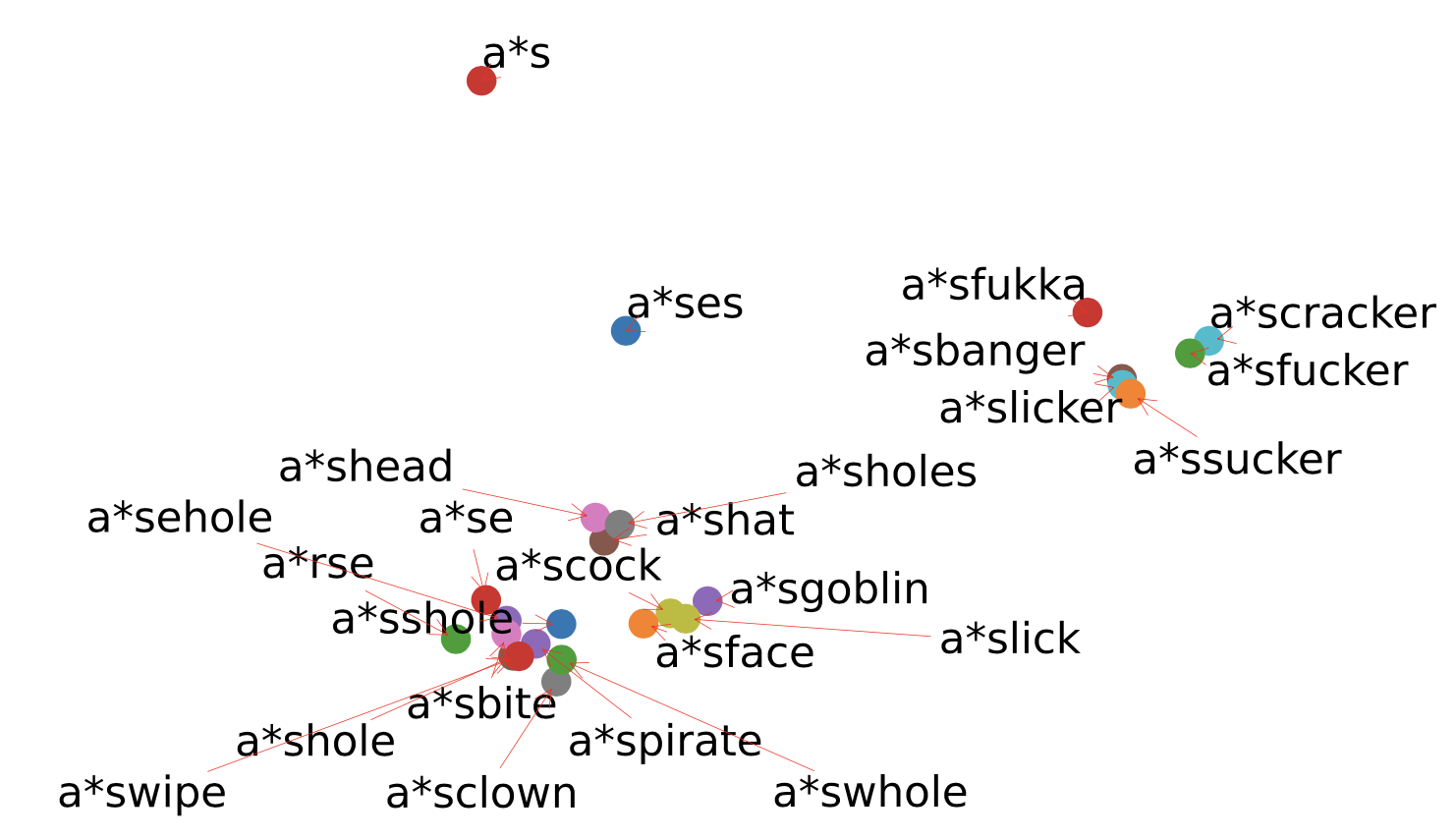}
    \caption{Cluster 1 of English abuses}
    \label{cluster_1}
\end{figure}

\begin{figure}[h]
    \centering
    \includegraphics[width=3.0 in]{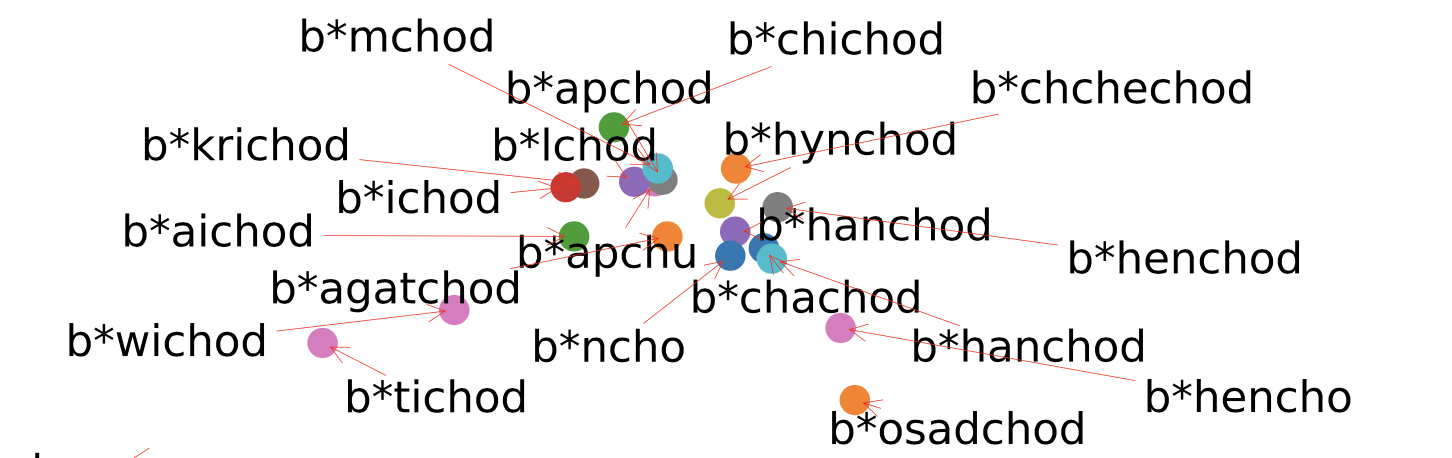}
    \caption{Cluster 2 of Hinglish abuses}
    \label{cluster_2}
\end{figure}

In order to better understand the learnt latent space of the model, we use the UMAP algorithm \cite{UMAP} to plot the word vectors in 2D space Fig. \ref{cluster_1} and \ref{cluster_2}. The figures clearly show that YZR-Net leverages the syntactic structure signal to cluster words and the first character is key in achieving that goal. Also in many cases the length of the word matters a lot in placing 2 tokens close to each other. This kind of structure suits our problem statement because we are not genuinely concerned about the semantics of the abuses, rather we care about how robustly our model can match the profane tokens or not.

\subsection{Deployment Architecture}
\begin{figure}[h]
    \centering
    \includegraphics[width=5.0 in]{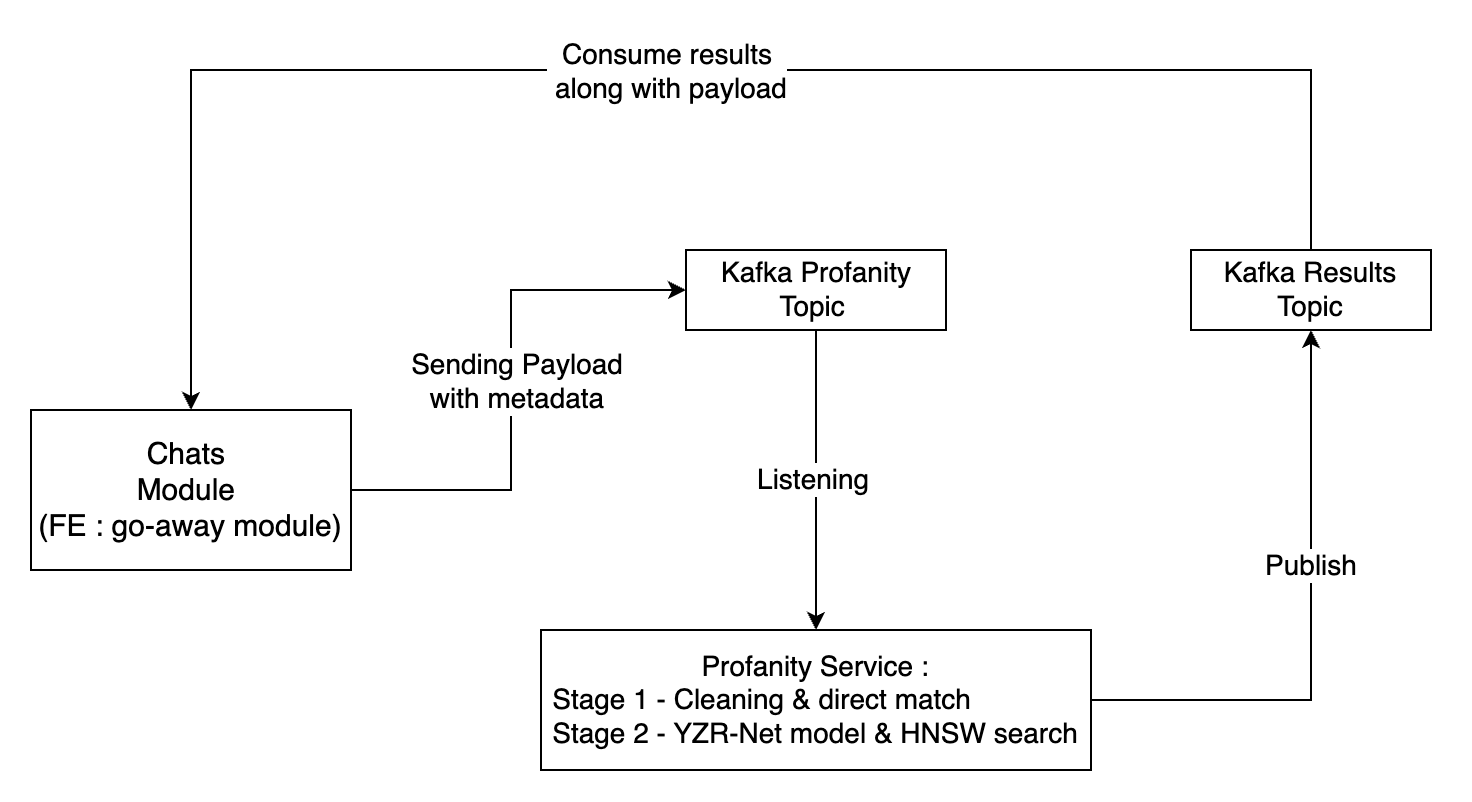}
    \caption{Deployment Architecture}
    \label{da}
\end{figure}

Another important challenge to be solved was setting up the entire architecture for the YZR-Net pipeline service consumption by our platform's chats framework. Due to the amount of data generated on a daily basis, a Kafka based setup was required to process the chats in an asynchronous manner. Fig \ref{da}, explains each and every module present in our architecture to process the chats in an efficient manner.

The processing flow is as follows :
\begin{itemize}
    \item \textbf{Chats Module} : The students post their chats inside this module, where we perform a light-weight profanity search in-case of direct matches. If there is no match, we send the chats along with meta-data to uniquely identify it, as a payload.
    \item \textbf{Profanity Topic} : All the chats marked as not profane by the light weight module, are published in the kafka topic of profanity for consumption by the next module in this pipeline.
    \item \textbf{Profanity Service} : This module is our YZR-Net pipeline which is divided into 2 stages :
        \begin{itemize}
            \item Stage 1 : Advanced matching \& suspicious token detection.
            \item Stage 2 : YZR-Net embedding generations \& HNSW search of suspicious tokens.
        \end{itemize}
    \item \textbf{Results Topic} : Once our services generates its predictions for a given chat, those results are published in the kafka topic of results along with original metadata present in its payload. This topic is used by our chats module to block profane messages and also used by our validation database to check whether our model needs to perform strict or lenient checking of chats in various situations. 
\end{itemize}

The most recent performance measurement of the aforementioned architecture on our daily chats data sample of 4229 chats contains 3968 not profane words (approximately 94\% of the data) and 261 profane words (approximately 6\% of the data). In this profane words, only 9 words were sent to the model and detected all 9 accurately. The remaining 252 words were identified by the proposed regex pipeline. 

% Thus, the developed model 

% gave us the following results :
% \begin{itemize}
%     \item Predicted NOT Profane :  ( 3968 / 4229 ) = \textbf{93.8\%}
%     \item Predicted Profane     :  ( 261  / 4229 ) = \textbf{6.17\%}
%         \begin{itemize}
%             \item Regular expression matches : (252 / 261) = \textbf{96.5\%}
%             \item YZR-Net pipeline match     : ( 9  / 261) = \textbf{3.44\%}
%         \end{itemize}
% \end{itemize}

On manual validation of results all the chats predicted as NOT profane were correct thereby giving us a precision of 1.0 for that class. For the chats predicted as PROFANE by direct regular expression match have to be correct but in many cases due to absence of context the predictions were incorrect from a student's point of view. Many terms used in Biology class are considered as profane on a day to day basis, therefore we had to modify our profane vocabulary based on the class at hand.

As for the 9 predictions made by our YZR-Net, 7 of them were absolutely correct but the remaining 2 predicted as PROFANE, were caused due to the spelling mistake made by the child. As per the model objective, YZR-Net did the right thing and there are no provisions to handle the spelling mistake scenarios.

\section{Conclusions and Future works}
We have presented a self-supervised based approach for detecting profane words in the user-generated text at e-learning platforms. To this end, we developed YZRnet based on SimCLR technique and investigate the effect of encoding customized augmentations to the default augmentations. Our best model achieves state-of-the-art performance in both automatic evaluations and human evaluations. Here we point out several interesting future research directions. Currently, our  model does not achieve the best performance across all variances (noise samples). We would like to explore how to better use the representations learned to improve the performance of profanity matching of all categories like various subjects, compound words etc. Besides this, it would also be interesting to consider to incorporate zero-shot learning in our model to further eliminate the human interventions for out-of-distribution samples. 

\bibliographystyle{plain}
\bibliography{profanity_references}
\end{document}